\documentclass{article}
\usepackage{spconf,amsmath,graphicx}
\usepackage{graphicx}
\usepackage{amsmath,amsfonts,amssymb}
\usepackage{mathtools}
\usepackage[acronym, toc, nonumberlist]{glossaries}
\usepackage[expproduct=times]{siunitx}
\usepackage{float}
\usepackage{pgfplots}
\usepackage{tikz}
\usepackage{caption}
\usepackage{subcaption}
\usepgfplotslibrary{groupplots}
\usetikzlibrary{arrows}
\usetikzlibrary{backgrounds}
\usetikzlibrary{fit}
\usetikzlibrary{positioning}
\usetikzlibrary{calc}
\usetikzlibrary{shapes.multipart}
\usepackage{balance}
\usepackage{url}

\usepackage{array}
\newcolumntype{L}[1]{>{\raggedright\let\newline\\\arraybackslash\hspace{0pt}}m{#1}}
\newcolumntype{C}[1]{>{\centering\let\newline\\\arraybackslash\hspace{0pt}}m{#1}}
\newcolumntype{R}[1]{>{\raggedleft\let\newline\\\arraybackslash\hspace{0pt}}m{#1}}

\usepackage{tabularx, etoolbox, booktabs}
\usepackage{multirow} 
\newcommand{\rarray}[1]{\renewcommand{\arraystretch}{#1}} 
\makeatletter
\def\zapcolorreset{\let\reset@color\relax\ignorespaces}
\def\colorrows#1{\noalign{\aftergroup\zapcolorreset#1}\ignorespaces}
\makeatother

\usepackage{algorithm}
\usepackage[noend]{algpseudocode}

\usepackage{xcolor}
\definecolor{yellow}{RGB}{255, 198, 0}
\definecolor{orange}{RGB}{255, 130, 0}
\definecolor{blue}{RGB}{0, 32, 91}
\definecolor{red}{RGB}{198, 53, 39}
\definecolor{magenta}{RGB}{138, 27, 97}
\definecolor{lightblue}{RGB}{0, 159, 223}
\definecolor{green}{RGB}{0, 155,119}
\definecolor{lightgreen}{RGB}{132, 189,0}
\definecolor{greenyellow}{RGB}{208, 223,0}

\newacronym{PA}{PA}{permutation alignment}
\newacronym{IBM}{IBM}{ideal binary mask}
\newacronym{DFT}{DFT}{discrete Fourier transformation}
\newacronym{STD}{STD}{standard deviation}
\newacronym{WSJ}{WSJ}{Wall Street Journal}
\newacronym{DNN}{DNN}{deep neural network}
\newacronym{TV-cGMM}{TV-cGMM}{time-variant complex Gaussian mixture model}
\newacronym{GMM}{GMM}{Gaussian mixture model}
\newacronym{vMFMM}{vMFMM}{von-Mises-Fisher mixture model}
\newacronym{vMF}{vMF}{von-Mises-Fisher}
\newacronym{cACGMM}{cACGMM}{complex angular-central Gaussian mixture model}
\newacronym{cWMM}{cWMM}{complex Watson mixture model}
\newacronym{cBMM}{cBMM}{complex Bingham mixture model}
\newacronym{EM}{EM}{expectation maximization}
\newacronym{DC}{DC}{deep clustering}
\newacronym{DAN}{DAN}{deep attractor network}
\newacronym{PIT}{PIT}{permutation invariant training}
\newacronym{GEV}{GEV}{generalized eigenvalue}
\newacronym{PSD}{PSD}{power spectral density}
\newacronym{SDR}{SDR}{signal to distortion ratio}
\newacronym{STFT}{STFT}{short time Fourier transform}
\newacronym{BLSTM}{BLSTM}{bidirectional long short-term memory network}
\newacronym{FF}{FF}{feed-forward}
\newacronym{AM}{AM}{acoustic model}
\newacronym{ASR}{ASR}{automatic speech recognition}
\newacronym{BAN}{BAN}{blind analytic normalization}
\newacronym{WER}{WER}{word error rate}
\newacronym{MVDR}{MVDR}{minimum variance distortionless response}
\newacronym{SNR}{SNR}{signal to noise ratio}
\newacronym{PESQ}{PESQ}{perceptual evaluation of speech quality}
\newacronym{NMF}{NMF}{non-negative matrix factorization}

\DeclareMathOperator{\trace}{tr}

\tikzset{
    every text node part/.style={align=center},
    >=stealth,
    block/.style={rectangle,draw=black!100,fill=gray!25,thick,minimum width=4.5em, minimum height=2em, inner sep=-2em,text depth=.25ex},
    branch/.style={circle,fill=black,minimum size=0.75mm,inner sep=0pt},
    random/.style={circle,draw=black!100,fill=lightgreen!50,thick,minimum size=2em,text height=1.5ex,text depth=.25ex},
    observation/.style={random, right=0.75cm of mask,double=lightgreen!50},
    parameter/.style={rectangle,draw=black!100,fill=red!50,thick,minimum size=8mm,text height=1.5ex,text depth=.25ex},
    arrow/.style={->,shorten >=0.05cm},
    reverse arrow/.style={<-,shorten <=0.05cm},
    apply/.style={circle,draw,fill=gray!50, minimum size=2.5mm,inner sep=0pt, label=center:{.}},
    node distance=0.5cm
}

\title{UNSUPERVISED TRAINING OF A DEEP CLUSTERING MODEL \\ FOR MULTICHANNEL BLIND SOURCE SEPARATION}
\name{Lukas Drude, Daniel Hasenklever, Reinhold Haeb-Umbach\vspace{-2mm}}
\address{Paderborn University, Department of Communications Engineering, Paderborn, Germany \\
\small\texttt{\{drude, haeb\}@nt.upb.de}\vspace{-4mm}}
\begin{document}
%
\maketitle
\begin{abstract}
We propose a training scheme to train neural network-based source separation algorithms from scratch when parallel clean data is unavailable.
In particular, we demonstrate that an unsupervised spatial clustering algorithm is sufficient to guide the training of a deep clustering system.
We argue that previous work on deep clustering requires strong supervision and elaborate on why this is a limitation.
We demonstrate that (a) the single-channel deep clustering system trained according to the proposed scheme alone is able to achieve a similar performance as the multi-channel teacher in terms of word error rates and (b) initializing the spatial clustering approach with the deep clustering result yields a relative word error rate reduction of \SI{26}{\percent} over the unsupervised teacher.
\end{abstract}
\section{Introduction}
\label{sec:intro}
In recent years, neural network architectures and training recipes demonstrated unprecedented performance in a wide range of applications.
In particular, \gls{DC} and \gls{PIT} are pioneering approaches to separate unseen speakers in a single channel mixture~\cite{Hershey2016DeepClustering, Kolbek2017Invariant}.
A plethora of subsequent work refined the architectures and training recipes, e.g. with respect to signal reconstruction performance by using a more direct reconstruction loss~\cite{Chen2017Attractor}, using multiple complementary losses~\cite{Luo2017StrongerTogether} or by addressing phase reconstruction~\cite{Roux2018Phasebook}.

In today's applications, e.g. digital home assistants/ meeting assistants, multiple microphones are the de-facto standard.
Thus, it is natural to generalize or apply neural network-based source separation to multi-channel scenarios.
Earlier work simply exploited spatial information for beamforming to extract the sources as a final processing step but did not make any use of spatial information to improve the clustering performance itself~\cite{Higuchi2017BF}.
Subsequently, two rather different approaches emerged, which made direct use of both spatial and spectral features:
On the one hand, directly feeding multi-channel features to a neural network yielded great improvements even for more than two channels~\cite{Wang2018MCDC}.
On the other hand, the embedding vectors stemming from either \gls{DC} or a \gls{DAN} can be seen as additional spectral features for an integrated \gls{EM} algorithm which jointly models spatial and spectral features in a single probabilistic model~\cite{Drude2017Integration}.

However, all of the aforementioned approaches need access to parallel clean data, i.e. recordings of the speech source or speech image at the microphones before any mixing took place.
Although it may seem that mixing signals artificially is sufficient for training, recent work with the CHiME 5 challenge dataset~\cite{Barker2018Chime} turned out to be surprisingly complicated:
Quasi-parallel data from the in-ear microphones was insufficiently related to the array recordings such that directly training a source separation neural network was impractical.
In particular, a correct transmission model of the room, the Lombard effect, and realistic background noise is hard to entirely simulate artificially.
One way may be to train an acoustic model with a neural network source separation system end-to-end.
However, this again requires transcriptions for the mixtures at hand~\cite{Yu2017PITE2E, Seki2018E2E} and may require pretraining the individual components using again parallel data~\cite{Chen2017Progressive, Settle2018E2E}.

Therefore, we here propose to train a source separation neural network from scratch by leveraging only spatial cues already during training.
This can be seen as a student-teacher approach~\cite{Bucila2006Compression}, where the unsupervised teacher turns out to be weaker than the student.
We demonstrate that a rough unsupervised spatial clustering approach is sufficient to guide the training of the neural network.

\section{Signal Model}
\label{sec:signal}
A convolutive mixture of $K$ independent source signals $s_{tfk}$, captured by $D$ sensors is approximated in the \gls{STFT} domain:
\begin{align}
\mathbf y_{tf}
= \sum\limits_k \mathbf h_{fk} \, s_{tfk} + \mathbf n_{tf}
= \sum\limits_k \mathbf x_{tfk} + \mathbf n_{tf}, \\[-7mm] \nonumber
\end{align}
where $\mathbf y_{tf}$, $\mathbf h_{fk}$ $\mathbf n_{tf}$, and $\mathbf x_{tfk}$ are the $D$-dimensional observed signal vector, the unknown acoustic transfer function vector of source $k$, the noise vector, and the source images at the sensors, respectively.
Furthermore, $t$ and $f$ specify the time frame index and the frequency bin index, respectively.
Since speech signals are sparse in the \gls{STFT} domain, we may assume that a time frequency slot is occupied either by a single source and noise or by noise only.

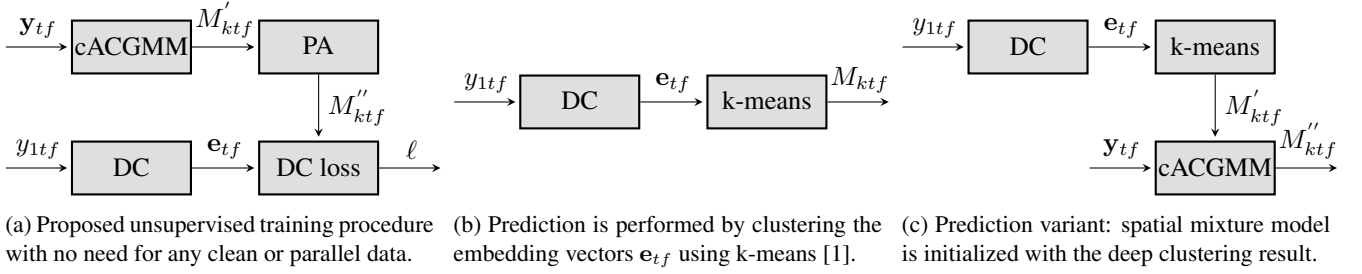
\begin{figure*}[t]
\centering
\subcaptionbox{Proposed unsupervised training procedure with no need for any clean or parallel data.}[0.32\textwidth]{\begin{tikzpicture}[
node distance=2.5em,
every text node part/.style={align=center}
]
\node (cacgmm) [block] {\acrshort{cACGMM}};
\node (pa) [block, right=of cacgmm] {\acrshort{PA}};
\node (blstm) [block, below=of cacgmm] {\acrshort{DC}};
\node (loss) [block, right=of blstm] {\acrshort{DC} loss};

\draw [reverse arrow] (cacgmm) edge node [above] {$\mathbf y_{tf}$} +(-4.75em, 0);
\draw [reverse arrow] (blstm) edge node [above] {$y_{1tf}$} +(-4.75em, 0);
\draw [arrow] (cacgmm) edge node [above] {$M_{ktf}^{'}$} (pa);
\draw [arrow] (blstm) edge node [above] {$\mathbf e_{tf}$} (loss);
\draw [arrow] (pa) edge node [right] {$M_{ktf}^{''}$} (loss);
\draw [arrow] (loss) edge node [above] {$\ell$} +(4.75em, 0);
\end{tikzpicture}
}%
\hspace{0.5em}
\subcaptionbox{Prediction is performed by clustering the embedding vectors $\mathbf e_{tf}$ using k-means~\cite{Hershey2016DeepClustering}.}[0.32\textwidth]{\raisebox{2.5em}{\begin{tikzpicture}[
node distance=2.5em,
every text node part/.style={align=center}
]
\node (blstm) [block] {\acrshort{DC}};
\node (kmeans) [block, right=of blstm] {k-means};

\draw [reverse arrow] (blstm) edge node [above] {$y_{1tf}$} +(-4.75em, 0);
\draw [arrow] (blstm) edge node [above] {$\mathbf e_{tf}$} (kmeans);
\draw [arrow] (kmeans) edge node [above] {$M_{ktf}$} +(4.75em, 0);
\end{tikzpicture}
}}%
\hspace{0.5em}
\subcaptionbox{Prediction variant: spatial mixture model is initialized with the deep clustering result.}[0.32\textwidth]{\begin{tikzpicture}[
node distance=2.5em,
every text node part/.style={align=center}
]
\node (blstm) [block] {\acrshort{DC}};
\node (kmeans) [block, right=of blstm] {k-means};
\node (cacgmm) [block, below=of kmeans] {\acrshort{cACGMM}};

\draw [reverse arrow] (blstm) edge node [above] {$y_{1tf}$} +(-4.75em, 0);
\draw [arrow] (blstm) edge node [above] {$\mathbf e_{tf}$} (kmeans);
\draw [arrow] (kmeans) edge node [right] {$M_{ktf}^{'}$} (cacgmm);
\draw [arrow] (cacgmm) edge node [above] {$M_{ktf}^{''}$} +(4.75em, 0);
\draw [reverse arrow] (cacgmm) edge node [above] {$\mathbf y_{tf}$} +(-4.75em, 0);
\end{tikzpicture}
}%
\caption{A deep clustering neural network is trained without parallel data by providing sufficient guidance by an unsupervised spatial clustering algorithm.
The resulting masks can then be used for beamforming or to directly mask the observed signal.
}
\label{fig:overview}
\end{figure*}

\section{Proposed framework}
We propose to use an unsupervised spatial clustering approach as a teacher for a neural network-based source separation student, thus rendering the whole setup to be unsupervised.
Subsec.~\ref{ssec:cacgmm} introduces the \gls{cACGMM} as an instance of an unsupervised spatial clustering algorithm and names inherent limitations.
Subsec.~\ref{ssec:dc} explains \gls{DC} as an example for a neural network-based source separation system which will then serve as the student.
Lastly, Subsec.~\ref{ssec:training} and \ref{ssec:prediction} detail, how training and prediction are performed.

\subsection{Unsupervised spatial clustering}
\label{ssec:cacgmm}

We here decided to use a \gls{cACGMM} for unsupervised spatial clustering~\cite{Ito2016cACGMM}, where the normalized complex-valued observation vectors $\mathbf{\tilde y}_{tf} = \mathbf y_{tf} / \Vert \mathbf y_{tf} \Vert$ are modeled as follows:
\begin{align*}
p(\mathbf{\tilde y}_{tf}; \boldsymbol\theta) = \sum_k \pi_{kf} \mathrm{cACG}(\mathbf{\tilde y}_{tf}, \mathbf B_{kf})
\end{align*}
with the model parameters $\boldsymbol\theta = \left\{\pi_{kf}, \mathbf B_{kf} \forall k,f\right\}$.
The observation model is a complex angular central Gaussian:
\begin{align}
\mathrm{cACG}(\mathbf{\tilde y}_{tf}, \mathbf B_{kf})
&= \frac{(D-1)!}{2\pi^{D}\det \mathbf B_{kf}} \frac{1}{\left(\mathbf{\tilde y}_{tf}^{\mathsf H} \mathbf B_{kf}^{-1} \mathbf{\tilde y}_{tf}\right)^{D}}.
\end{align}

All latent variables and parameters can be estimated on each mixture signal using an \gls{EM} algorithm\textsuperscript{\ref{pb_bss}}.
The \gls{EM} update equations coincide with the update equations of the \gls{TV-cGMM}~\cite{Ito2014Clustering} as demonstrated in the appendix of~\cite{Ito2016cACGMM}.
The obtained class affiliation posteriors can then be used to extract the sources either by masking or beamforming.

However, it is worth noting that mixture models, in general, tend to be very susceptible to initialization.
This will be addressed in more detail in the evaluation section.

The \gls{cACGMM} neglects frequency dependencies.
Thus, when used without any kind of guidance, it will yield a solution where the speaker index is inconsistent over frequency bins.
This issue is the so-called frequency permutation problem~\cite{Sawada2007Permutation}.
We address it by calculating that \gls{PA} (bin by bin) which maximizes the correlation of the masks along neighboring frequencies similar to~\cite{Sawada2007Permutation}\footnote{Our implementation of the \gls{cACGMM} and \acrlong{PA} can be found on Github: \url{https://github.com/fgnt/pb_bss}\label{pb_bss}}.

\subsection{Deep Clustering}
\label{ssec:dc}

\Gls{DC} is a technique which aims at blindly separating unseen speakers in a single-channel mixture.
The training procedure described in the original work~\cite{Hershey2016DeepClustering, Isik2016DeepClustering} assumes that \glspl{IBM} for each speaker are available to train a multi-layer \gls{BLSTM}~\cite{Schuster1997BLSTM} to map from $T \cdot F$ spectral features (e.g. log spectrum) to the same number of $E$-dimensional embedding vectors $\mathbf e_{tf}$, where $\Vert \mathbf e_{tf} \Vert^2 = 1$.
The objective during training is to minimize the Frobenius norm of the difference between the estimated and true affinity matrix (for a discussion of improved loss functions see ~\cite{Wang2018Alternative}):
\begin{align}
\ell = \left\Vert\smash{\mathbf{\hat A}} - \mathbf A^{\vphantom{\mathsf T}}\right\Vert^2_{\mathrm F}
= \left\Vert\mathbf E \mathbf E^{\mathsf T} - \mathbf C \mathbf C^{\mathsf T}\right\Vert^2_{\mathrm F},\label{dc_loss}
\end{align}
where $\mathbf{\hat A}$ and $\mathbf A$ are the estimated and ground truth affinity matrices.
The entries $A_{n,n'}$ encode, whether observation $n$ and $n'$ belong to the same source ($A_{n,n'}=1$, and zero else).
Correspondingly, the embeddings are stacked in a single matrix $\mathbf E$ with shape $(TF \times E)$ and the ground truth one-hot vectors describing which time frequency slot belongs to which source are stacked in a single matrix $\mathbf C$ with shape $(TF \times K)$, such that $C_{nk} = 1$, if observation $n$ belongs to source $k$ and $C_{nk} = 0$ otherwise.

During training, the network is encouraged to move embeddings belonging to the same source closer together while pushing embeddings which belong to different sources further apart.
After training, the embeddings, which are normalized to unit-length, can be clustered to obtain time frequency masks for each source.
The original work used k-means clustering.
This yields masks which can be used in a subsequent source extraction scheme e.g. masking or beamforming.
\vfill

\subsection{Unsupervised network training}
\label{ssec:training}

To train a \gls{DC} system, we need to optimize the affinity loss in Eq.~\ref{dc_loss}.
Since ground truth is not available, we instead minimize the Frobenius norm of the difference between the affinity matrix of the embeddings $\mathbf E \mathbf E^{\mathsf T}$ and the affinity matrix according to masks predicted by an unsupervised clustering approach.
Here, we use the class predictions (masks) $M_{ktf}^{'}$ of a \gls{cACGMM} (Subsec.~\ref{ssec:cacgmm}) and apply an additional frequency permutation alignment (Subsec.~\ref{ssec:cacgmm}) to obtain $M_{ktf}^{''}$ (Fig.~\ref{fig:overview}~(a)), which is then used to guide the optimization.

\begin{figure}[b]
\vspace{-2mm}
\newlength\figureheight
\newlength\figurewidth
\setlength\figureheight{5.8cm}
\setlength\figurewidth{4.85cm}
\begin{tikzpicture}
\tikzstyle{every node}=[font=\small]
\pgfplotsset{compat=newest}
\begin{groupplot}[group style={group size=2 by 1, horizontal sep=0.2cm, vertical sep=0.4cm}]
\nextgroupplot[
xlabel={Time frame index},
x label style={yshift=1mm},
xmin=-0.5, xmax=740.5,
ymin=-0.5, ymax=256.5,
xtick={-100,0,360,700},
xticklabels={-100,0,200,400},
ytick={0,250},
yticklabels={0,257},
ylabel={Frequency bin index},
y label style={yshift=-4mm},
width=\figurewidth,
height=\figureheight,
tick align=inside,
tick pos=left,
x grid style={white!69.01960784313725!black},
y grid style={white!69.01960784313725!black},
trim axis left,
trim axis right
]
\addplot graphics [includegraphics cmd=\pgfimage,xmin=-0.5, xmax=740.5, ymin=-0.5, ymax=256.5] {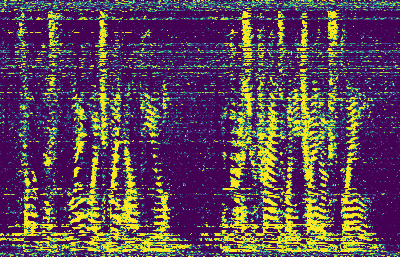};

\nextgroupplot[
colorbar,
colorbar style={xshift=-2.5mm, ytick={0, 1},yticklabels={0.0, 1.0}, ylabel={Posterior mask $\gamma_{1tf}$}, width=0.15cm, xshift=0mm, ylabel style={yshift=4mm}},
colormap/viridis,
xlabel={Time frame index},
x label style={yshift=1mm},
ymajorticks=false,
xmin=-0.5, xmax=740.5,
ymin=-0.5, ymax=256.5,
xtick={-100,0,360,700},
xticklabels={-100,0,200,400},
minor tick num=4,
width=\figurewidth,
height=\figureheight,
tick align=inside,
tick pos=left,
x grid style={white!69.01960784313725!black},
y grid style={white!69.01960784313725!black}
trim axis left,
trim axis right
]
\addplot graphics [includegraphics cmd=\pgfimage,xmin=-0.5, xmax=740.5, ymin=-0.5, ymax=256.5] {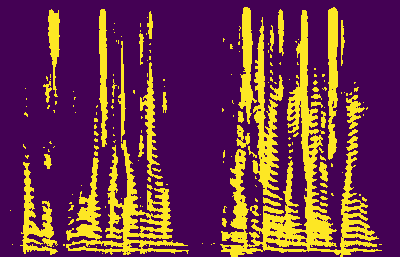};

\end{groupplot}

\end{tikzpicture}
\setlength{\belowcaptionskip}{-3pt}
\vspace{-1em}
\caption{
Intermediate masks generated by the \gls{cACGMM} (left) guide the neural network training which results in k-means masks with less artifacts (right).
Especially the lower frequencies are resolved better.
}
\label{fig:masks}
\end{figure}

It is worth noting that the predicted classes (masks) from the spatial mixture model are first of all less sharp than the unavailable \glspl{IBM}.
Furthermore, the predicted masks often contain frequency permutations errors, when the permutation alignment step did not resolve all permutations (see Fig.~\ref{fig:masks}, left).
The assumption which has to be proven in the evaluation section, therefore, is that on average, the \gls{cACGMM} results are sufficiently good.

Since any kind of ground-truth signal is not available in the unsupervised case, early stopping~\cite{prechelt1998automatic} can only be performed with respect to the aforementioned loss and not with respect to \gls{SDR} gains or similar.

\subsection{Prediction}
\label{ssec:prediction}
To predict masks at test time two different ways are natural to investigate.
First, one can use the trained \gls{DC} network to predict embeddings $\mathbf e_{tf}$ which are then clustered using k-means as in Fig.~\ref{fig:overview} (b).
This approach is the recommended approach according to~\cite{Hershey2016DeepClustering}.
Second, one may use the k-means masks as initialization for a \gls{cACGMM} as in Fig.~\ref{fig:overview} (c).
This can be seen as some kind of weak integration (in contrast to~\cite{Drude2017Integration}) between the \gls{DC} model, which just uses spectral features and the \gls{cACGMM} which just uses spatial features due to the normalization as mentioned in Subsec.~\ref{ssec:cacgmm}.

\section{Beamforming}
\label{sec:bf}
Both aforementioned models provide a time-frequency mask for each speaker and noise.
This can be multiplied with the observed signal \gls{STFT} to extract each of the sources.
Alternatively, we may use it for classic statistical beamforming as a linear time-invariant way to extract each source.
To do so, we first calculate covariance matrices for each target speaker:\vspace{-0.5em}
\begin{align}
\boldsymbol\Phi_{kf}^{\mathrm{(target)}}
&= \left. \sum_t M_{ktf}^{\mathrm{(target)}} \mathbf y_{tf} \mathbf y_{tf}^{\mathsf H} \middle/
\sum_t M_{ktf}^{\mathrm{(target)}} \right..
\end{align}
Similarly, the covariance matrix of all interferences and noise is calculated with $\smash{M_{ktf}^{\mathrm{(inter)}} = 1 - M_{ktf}^{\mathrm{(target)}}}$.

These covariance matrices can then be used to derive a statistically optimal beamformer.
In this particular case, we opted for a specific formulation of the \gls{MVDR} beamformer which avoids explicit knowledge of any kind of steering vector~\cite[Eq.~24]{Souden2010MVDR}:
\begin{align}
\mathbf w
&= \frac{1}{\lambda_{kf}} \boldsymbol \Phi_{kf} \mathbf u_k, ~~\text{with}~~
\boldsymbol \Phi_{kf} = {\boldsymbol\Phi_{kf}^{\mathrm{(inter)}}}^{-1} \boldsymbol\Phi_{kf}^{\mathrm{(target)}},
\end{align}
where $\lambda_{kf} = \trace\left(\boldsymbol \Phi_{kf}\right)$ and $\mathbf u_k = [0 ... 1 ... 0]^{\mathsf T}$ selects the reference microphone.
Here, the reference microphone is selected by maximizing the expected \gls{SNR} gain~\cite{Erdogan2016MVDR}.
The beamforming vector is then used to linearly project the multi-channel mixture to a single-channel estimate of clean speech: $z_{ktf} = \smash{\mathbf w_{kf}^{\mathsf H} \mathbf y_{tf}}$.

\section{Acoustic model}
\label{sec:am}
For an objective comparison, we train a state of the art \gls{AM}.
The hybrid \gls{AM} consists of a combination of a wide residual network to model local context and a \gls{BLSTM} to model long term dependencies.
The \gls{AM} is thus dubbed wide bi-directional residual network~\cite{Heymann2016WBRN}.
To train the \gls{AM} alignments are extracted with a vanilla DNN-HMM recipe from Kaldi~\cite{Povey2011Kaldi}.
The \gls{AM} is trained on artificially reverberated \gls{WSJ} utterances without any interfering speaker.
Thus, the \gls{AM} never saw mixed speech and never saw possible artifacts produced by any kind of separation system.
For decoding, we use the \gls{WSJ} trigram language model without additional rescoring.

\section{Evaluation}
\label{sec:evaluation}
\setlength{\dbltextfloatsep}{8.0pt plus 2.0pt minus 2.0pt}
\begin{table*}[tb!]
	\rarray{0.98}
	\caption{
    	Model comparison for unsupervised systems in terms of objective quality gains and \glspl{WER}.
    	Oracle initialized baseline and supervised baseline are denoted in gray.
        The best unsupervised result is set in bold.
	}
	\label{table:result}
	\robustify\bfseries 
	\centering
    \newcommand{\mrow}[2]{\multirow{#1}{*}[-0.85pt]{#2}}
	\begin{tabular}{r@{\hskip 2mm}l@{\hskip 2mm}llc@{\hskip 2mm}cc@{\hskip 2mm}c@{\hskip 2mm}c@{\hskip 2mm}c@{\hskip 2mm}c}
		\toprule
        & Model & Initialization\!\!\!\! & Extraction\!\!\!\! &
        \multicolumn{2}{c}{BSS-Eval SDR gain} &  \multicolumn{2}{c}{Invasive SDR gain} & \!\!PESQ gain & STOI gain\! & WER / \si{\percent} \\
        \cmidrule(lr){5-6} \cmidrule(lr){7-8}
        & & & & Mean / \si{dB} & STD / \si{dB} & Mean / \si{dB} & STD / \si{dB} & & & \\
        \midrule
                            1 &                    cACGMM &                         Random &                    Masking &                    7.2 &                             2.8 &                    10.4 &                    3.3 &                     0.17 &                    0.11 &                    38.4 \\
                    2 &                      U-DC &                         Random &                    Masking &                    5.5 &                             3.9 &                     9.4 &                    3.7 &                    -0.42 &                    0.04 &                    75.1 \\
                    3 &                    cACGMM &                    U-DC result &                    Masking &           \textbf{9.5} &                    \textbf{2.4} &                    13.2 &           \textbf{2.9} &                     0.40 &           \textbf{0.18} &                    29.3 \\
  \textcolor{gray}{4} &    \textcolor{gray}{S-DC} &       \textcolor{gray}{Random} &  \textcolor{gray}{Masking} &  \textcolor{gray}{5.9} &           \textcolor{gray}{4.8} &   \textcolor{gray}{9.5} &  \textcolor{gray}{3.4} &  \textcolor{gray}{-0.25} &  \textcolor{gray}{0.06} &  \textcolor{gray}{75.8} \\
  \textcolor{gray}{5} &  \textcolor{gray}{cACGMM} &  \textcolor{gray}{S-DC result} &  \textcolor{gray}{Masking} &  \textcolor{gray}{9.1} &  \textcolor{gray}{{2.4}} &  \textcolor{gray}{12.6} &  \textcolor{gray}{2.8} &   \textcolor{gray}{0.37} &  \textcolor{gray}{0.16} &  \textcolor{gray}{31.0} \\
  \textcolor{gray}{6} &  \textcolor{gray}{cACGMM} &   \textcolor{gray}{Oracle IBM} &  \textcolor{gray}{Masking} &  \textcolor{gray}{9.7} &           \textcolor{gray}{2.3} &  \textcolor{gray}{13.3} &  \textcolor{gray}{2.6} &   \textcolor{gray}{0.48} &  \textcolor{gray}{0.14} &  \textcolor{gray}{28.9} \\
\midrule
                    7 &                    cACGMM &                         Random &                       MVDR &                    5.1 &                             3.2 &                    12.7 &                    4.0 &                     0.37 &                    0.09 &                    28.0 \\
                    8 &                      U-DC &                         Random &                       MVDR &                    5.7 &                             3.5 &                    13.6 &                    4.5 &                     0.43 &                    0.11 &                    29.0 \\
                    9 &                    cACGMM &                    U-DC result &                       MVDR &                    6.4 &                             3.4 &           \textbf{15.3} &                    3.5 &            \textbf{0.52} &                    0.13 &           \textbf{20.7} \\
 \textcolor{gray}{10} &    \textcolor{gray}{S-DC} &       \textcolor{gray}{Random} &     \textcolor{gray}{MVDR} &  \textcolor{gray}{5.9} &           \textcolor{gray}{3.2} &  \textcolor{gray}{14.2} &  \textcolor{gray}{4.1} &   \textcolor{gray}{0.47} &  \textcolor{gray}{0.12} &  \textcolor{gray}{26.5} \\
 \textcolor{gray}{11} &  \textcolor{gray}{cACGMM} &  \textcolor{gray}{S-DC result} &     \textcolor{gray}{MVDR} &  \textcolor{gray}{6.1} &           \textcolor{gray}{3.3} &  \textcolor{gray}{14.9} &  \textcolor{gray}{3.5} &   \textcolor{gray}{0.50} &  \textcolor{gray}{0.12} &  \textcolor{gray}{21.1} \\
 \textcolor{gray}{12} &  \textcolor{gray}{cACGMM} &   \textcolor{gray}{Oracle IBM} &     \textcolor{gray}{MVDR} &  \textcolor{gray}{6.4} &           \textcolor{gray}{3.3} &  \textcolor{gray}{15.5} &  \textcolor{gray}{3.4} &   \textcolor{gray}{0.78} &  \textcolor{gray}{0.12} &  \textcolor{gray}{19.9} \\
		\bottomrule
	\end{tabular}
\end{table*}

To evaluate the proposed algorithm, we artificially generated 30000, 500 and 1500 six channel mixtures with a sampling rate of \SI{8}{kHz} with source signals obtained from three non-overlapping \gls{WSJ} sets (\texttt{train\_si284}, \texttt{cv\_dev93}, \texttt{test\_eval92}).
We generated room impulse responses with the image method~\cite{Allen1979Image} with random room dimensions, a random position of the circular array and random positions of the two concurring speakers.
The minimum angular distance was set to \SI{15}{\degree}.
The reverberation time (T60) was uniformly sampled between \num{200} and \SI{500}{ms}.
Additive white Gaussian noise with \num{20} to \SI{30}{dB} \gls{SNR} was added to the mixture.
The source separation algorithms operate on \gls{STFT} signals with a \gls{DFT} size of 512 and a shift of 128.
The \gls{DC} network consists of 2 \gls{BLSTM} layers with 600 forward and 600 backward units and a final linear layer which yields $E=20$ dimensional embeddings.
The \gls{AM} uses 40 Mel filterbank features extracted with an \gls{STFT} with a \gls{DFT} size of 256, a window size of 200 and a shift of 80.

To get a good impression of the system performance, we present results in terms of mean and \gls{STD} of BSS-Eval \gls{SDR} gain~\cite{Vincent2006Performance}, mean and \gls{STD} of invasive \gls{SDR} gain similar as in~\cite{TranVu2010Blind}, PESQ gain~\cite{Rix2001PESQ}, STOI gain~\cite{Taal2011STOI} and finally \glspl{WER}.
The invasive \gls{SDR} is the power ratio of applying the obtained mask or beamforming vector to the target speech image and the sum of all interference images.

First of all, we evaluated unsupervised spatial clustering with a \gls{cACGMM} using a random initialization.
Comparing row 1 and 7, we observe that although masking yields higher BSS-Eval \gls{SDR} and STOI gains, the invasive \gls{SDR} gain, \gls{PESQ} gain and foremost the \gls{WER} are better for the \gls{MVDR} beamforming approach.
This can be explained by the observation that masking often leads to musical tones which are avoided with a time-invariant projection approach such as beamforming.

Next, we trained the \gls{DC} system according to the proposed training scheme using masks exported according to row 1 or 7.
The performance of the unsupervisedly trained \gls{DC} (row 8, abbreviated as U-DC) system does not quite reach the performance of the supervised \gls{DC} in row 10 (S-DC).
Comparing the training times, the unsupervised approach takes at least twice as many training steps (here 500k steps, batch size 4) as the supervised \gls{DC} system.
The sometimes misleading \gls{cACGMM} masks which result in more noisy gradients are a possible explanation.

Rows 3 and 9 summarize the results obtained when initializing the \gls{cACGMM} with the U-DC result of row 2 or 8, respectively.
When using masking, this results in a gain of \SI{2.6}{dB} invasive \gls{SDR} over the randomly initialized \gls{cACGMM} in row 1.
Also, when using beamforming, a gain of \SI{2.6}{dB} invasive \gls{SDR} can be observed.
Comparing this with supervisedly trained \gls{DC} in row 10 and the S-DC initialized \gls{cACGMM} in row 11, it can be seen that the proposed approach outperforms both oracle baselines.
Interestingly, the initialization is sufficient to address the permutation problem such that an additional alignment step is not necessary anymore.
It is therefore valid to conclude that the main drawback of the teacher is its initialization.
However, the performance of a \gls{cACGMM} initialized with oracle ideal binary masks is not quite reached.
Overall, the spatial mixture model initialized with the unsupervised \gls{DC} (row 9) result yields a relative \gls{WER} reduction of \SI{26}{\percent} over the unsupervised teacher (row 7) and beats the best supervised system (row 11) by \SI{2}{\percent} relative.

\section{Conclusion}
\label{sec:conclusion}

In this work, we demonstrated that a neural network-based blind source separation system can indeed be trained from scratch without any clean or parallel data.
We opted for a training scheme with an unsupervised teacher and ended up with a student outperforming the teacher.
This is of particular interest when addressing real data such as the CHiME 5 challenge recordings, where regular training is impeded due to the lack of well-matching targets.
We see this work as an alternative/ complementary tool to train the frond-end entirely end-to-end or to better match real recordings and plan to extend the present work to CHiME 5 challenge recordings.

\clearpage
\ninept
\bibliographystyle{IEEEbib}
\bibliography{strings,refs}

\end{document}